%% file: paper.tex
\documentclass[11pt]{article}
\usepackage{acl}
\setcitestyle{numbers,square}
\usepackage{times}
\usepackage[T1]{fontenc}
\usepackage[utf8]{inputenc}
\usepackage{microtype}
\usepackage{amsmath,amssymb,amsthm,booktabs,graphicx,longtable,float}
\newtheorem{lemma}{Lemma}
\newtheorem{proposition}{Proposition}
\makeatletter
\newcommand{\Needspace}[1]{\par\penalty-100\begingroup\setlength{\dimen@}{#1}\dimen@ii\pagegoal\advance\dimen@ii-\pagetotal\ifdim\dimen@>\dimen@ii\ifdim\dimen@ii>\z@\vfil\fi\break\fi\endgroup}
\makeatother
\usepackage{lipsum}
\usepackage{enumitem}
\newcommand{\E}{\mathbb{E}}
\newcommand{\Var}{\operatorname{Var}}
\newcommand{\Beta}{\operatorname{Beta}}

\title{Few Trials, Many Metrics: Beta Mixtures for Low-Budget AI Evaluation}
\author{
  \textbf{Gustavo Tironi}\textsuperscript{1} \quad
  \textbf{Felipe Maia Polo}\textsuperscript{2} \quad
  \textbf{Diego Mesquita}\textsuperscript{1} \\
  \textsuperscript{1}Getulio Vargas Foundation \\
  \textsuperscript{2}Institute of Foundation Models, MBZUAI \\
  \texttt{gustavo.tironi@fgv.edu.br}
}

\begin{document}
\maketitle

\begin{abstract}
%Estimating how an AI system performs over repeated attempts, or how sensitive it is to evaluation choices, can require many model calls. Hierarchical latent variable models, such as the beta-binomial, have been used to estimate Pass@$k$ from fewer trials. We show that the same kind of model supports a wider range of evaluation tasks. From one fitted Beta mixture, we derive Pass$^k$ and AUC@$k$ as functions of the model parameters, which lets us estimate these metrics at large horizons $k$ from a few trials per task. We also derive confidence intervals for these metrics and for Pass@$k$ with the delta method, without extra model calls. The same model supports AI sensitivity analysis by treating configurations, such as prompt templates, as instances. We argue that hierarchical latent variable models can serve AI evaluation well beyond a single metric. We validate both applications on real benchmark data. For repeated success metrics, we found that the proposed approach
Evaluating a stochastic AI system often requires many repeated samples per instance, especially for metrics defined over long horizons. Recent work has shown that modeling instance-level success probabilities with a Beta distribution recovers Pass@$k$ from far fewer samples than direct estimation. We show that this framework supports a broader range of evaluation targets. We fit a mixture of Beta distributions by marginal maximum likelihood, with the number of components chosen by BIC. From that one fit, we compute Pass$^k$ and AUC@$k$ at any horizon in closed form. We also derive delta-method confidence intervals for these metrics and for Pass@$k$, which require no additional model calls. Treating configurations rather than tasks as instances extends the same model to sensitivity analysis. For repeated success metrics, we find that mixture priors yield more accurate estimators than naive estimation across competition mathematics and graduate-level science benchmarks when extrapolating beyond the observed number of trials, and the resulting intervals cover the full-data reference. When prompts are treated as instances, it estimates the distribution of prompt sensitivity more accurately than a comparable item-response baseline without external covariates. Direct estimators remain preferable once trials reach the target horizon.
\end{abstract}

\input{sections/intro}
\input{sections/related}

\input{sections/method}
\input{sections/experiments}
\input{sections/conclusion}

\nocite{dekoninck2026matharena}
\bibliography{paper}

\input{sections/appendix}
\end{document}

%% file: sections/intro.tex
\section{Introduction}

Statistical modeling plays a central role in how we evaluate artificial intelligence (AI) systems and analyze benchmark data. Classic tools from psychometrics and paired-comparison analysis, such as item response theory \citep{rasch1993probabilistic,lord1952theory,reckase200618} and the Bradley-Terry model \citep{zermelo1929berechnung, bradley1952rank}, use latent variables and model parameters to capture differences across both systems and evaluation instances \citep{rodriguez2021evaluation,polo2024tinybenchmarks,chiang2024chatbot}. These methods are mostly used to compare many systems against one another. More recent work takes a different approach and applies other hierarchical latent-variable models to repeated evaluations of a single, fixed system. The beta-binomial model, for example, has been used to estimate Pass@$k$ \citep{kazdan2025efficient,verine2026estimating} from fewer trials and to characterize overdispersion in benchmark outcomes \citep{manousaridis2026characterizing,buran2026benchmarks}. In this paper, we focus on this second line of work.

\begin{figure}[t]
 \centering
 \includegraphics[width=\columnwidth]{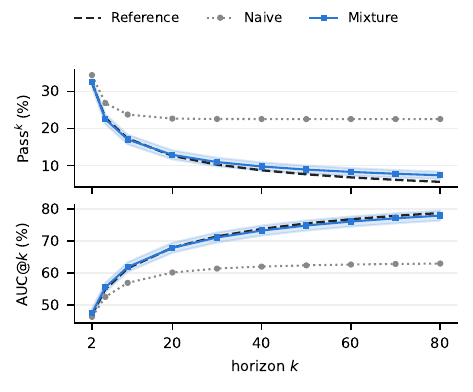}
 \caption{From 5 trials per task, Mixture tracks the 80-trial
 reference at every horizon. One GPQA group; band:
 95\% interval (Appendix~\ref{sec:teaser-details}).}
 \label{fig:teaser}
\end{figure}

Consider a fixed AI system evaluated on a collection of instances\footnote{An instance is the smallest unit of evaluation, such as a single question or user prompt.}. Because the system's responses are stochastic, we can query it $m_j$ times on the same instance $j$ and record an outcome $Y_j$ that summarizes these repeated attempts, such as the number of correct responses. In a hierarchical model such as the beta-binomial, each instance depends on a latent variable $\theta$ drawn from a shared distribution $P_\gamma$, and the outcome follows $Y_j \mid \theta \sim P(Y \mid \theta)$. Integrating out the latent variable gives the marginal distribution of an outcome,
\begin{equation*}\textstyle
    \mathbb{P}_\gamma(Y_j = y) = \int \mathbb{P}(Y_j = y \mid \theta)\, \mathrm{d}P_\gamma(\theta),
\end{equation*}
thus the behavior of the whole population of instances in a benchmark is governed by $\gamma\in\Gamma\subseteq\mathbb{R}^d$. This separates two sources of variation: randomness in the system's responses to a given instance, captured by $P(Y \mid \theta)$, and differences between instances, captured by $P_\gamma$. The beta-binomial model is the special case in which $\theta \in [0,1]$ is the probability that a single sampled response to instance $j$ is correct, $\theta$ follows a Beta distribution, and $Y_j \mid \theta \sim \mathrm{Binomial}(m_j, \theta)$ counts the correct responses among the $m_j$ samples.

We argue that this framework supports a broader range of evaluation objectives beyond estimating Pass@$k$. Once $P_\gamma$ has been estimated, other widely used metrics, such as Pass$^k$ \citep{yao2024taubench} and AUC@$k$, can be expressed as deterministic functions of $\gamma$ and estimated with a limited number of model calls. Leveraging maximum likelihood theory, we also derive rigorous asymptotic confidence intervals for these estimates, including Pass@$k$.

In another direction, we show that the same framework can also be used to study sources of variability beyond differences between tasks. One application we explore is \emph{AI sensitivity analysis}. By treating system configurations such as system prompts, instruction formats, and decoding hyperparameters as instances, we can estimate how much measured system performance varies across these choices, even on a limited budget.

%In another direction, we show that the same framework also lets us study sources of variability other than differences between tasks. One major application we explore is \emph{AI sensitivity analysis}. If we let configurations such as system prompts, instruction formats, and decoding hyperparameters play the role of instances, we can estimate how much a system's measured performance shifts across these choices, even under a limited evaluation budget.

To summarize, our contributions are threefold:
\vspace{-.225cm}
\begin{itemize}[leftmargin=*]
\item We develop low-budget estimators for Pass$^k$ and AUC@$k$, along with asymptotic confidence intervals for these metrics and Pass@$k$, enabling efficient estimation with quantified uncertainty.
 \looseness=-1
\vspace{-.25cm}
\item We extend the same idea to AI sensitivity analysis by treating configurations, such as prompt templates, as instances, estimating performance variability at a low-resource scenario.
\vspace{-.25cm}
\item We evaluate both use cases on real benchmark data, showing that our method outperforms competing approaches in accuracy while providing valuable insights under tighter budgets.
\end{itemize}

%\begin{itemize}[leftmargin=*]
%    \item We formulate widely used evaluation metrics (Pass$^k$ and AUC@$k$) as statistical estimands under a common hierarchical model, expressing each as a function of $\gamma$. We also derive asymptotic confidence intervals for these metrics and for Pass@$k$, so they can be estimated on a limited budget with quantified uncertainty. \looseness=-1
%    \vspace{-.25cm}
%    \item We apply the same hierarchical model to sensitivity analysis of AI systems. By treating evaluation configurations, such as prompt templates, as instances, the fitted population distribution estimates how much a system's performance varies across these choices, without evaluating every configuration in full.
%    \vspace{-.25cm}
%    \item We evaluate both use cases on real benchmark data and show that, from sparse observations, the typical group reaches an estimate within $3$ percentage points of the 80-trial reference with $78\%$ fewer trials ($4.6\times$ less compute). For prompt sensitivity, the mixture reduces the $W_1$ error of pIRT by $46\%$ on average across benchmarks and budgets, without external prompt or question features.
%\end{itemize}

%% file: sections/related.tex
\subsection{Related Work}
\label{sec:related}

\paragraph{Efficient AI evaluation.}
Many methods estimate full-benchmark performance from a small subset of items, either selecting items using reference models' past results~\citep{vivek2024anchor,perlitz2024efficient,polo2024tinybenchmarks,kipnis2025metabench} or acquiring them actively~\citep{kossen2021active,berrada2025scaling}. \citet{zhang2025benchmark} find these methods rarely beat a random-sample mean when extrapolating to stronger models, while \citet{cordero2026bayesame} show that Bayesian active selection can, and can also set the subset size automatically. Evaluation cost also comes from repeated sampling: the unbiased Pass@$k$ estimator needs at least $k$ generations per task~\citep{chen2021evaluating}. Rather than extrapolating power laws fit to small-$k$ estimates~\citep{brown2024monkeys,hughes2024bestofn}, modeling the distribution of task difficulty recovers Pass@$k$ curves from far fewer generations~\citep{levi2024simple,schaeffer2025monkeys,kazdan2025efficient,verine2026estimating}.

\paragraph{AI sensitivity analysis.} Prior work treats AI evaluation as a sensitivity analysis. Prompt formatting and instruction paraphrases can substantially shift both scores and model rankings~\citep{sclar2024quantifying,mizrahi2024state}. This has motivated efficient multi-prompt estimation~\citep{polo2024efficient}, large-scale perturbation datasets~\citep{habba-etal-2025-dove}, and analyses of the full evaluation system~\citep{tian2026beyond}.

%% file: sections/method.tex
\section{Hierarchical models for AI evaluation}
\label{sec:method}

We first define the evaluation targets and their direct estimators. We then describe a model for metric estimation and sensitivity analysis.

\subsection{Evaluation targets}
\label{sec:metrics}
For a fixed system, instance $j$ has success probability $\theta_j$ and $y_j$ successes in $m_j$ conditionally independent attempts. For $\theta\sim P_\gamma$, Pass$^k$ is the probability that all $k$ attempts succeed \citep{yao2024taubench}, while Pass@$k$ is the probability of at least one success, $\E_\gamma[1-(1-\theta)^k]$ \citep{chen2021evaluating}. We summarize the latter across horizons using AUC@$k$,
\begin{equation*}
\begin{aligned}
 \mathrm{Pass}^{k}&=\E_\gamma[\theta^k],\\
 \mathrm{AUC@}k&=k^{-1}\sum_{r=1}^{k}\mathrm{Pass@}r.
\end{aligned}
\end{equation*}
AUC averages every integer horizon through $k$. Both metrics equal accuracy $\E_\gamma[\theta]$ at $k=1$.

While tempting, using the naive maximum likelihood estimator for $\theta_j$, i.e.\ $y_j/m_j$, and averaging the resulting metric over instances yields upward bias for Pass$^k$ and downward bias for AUC@$k$ when $k>1$. Unbiased combinatorial estimators require $m_j\geq k$ \citep{chen2021evaluating,yao2024taubench}. We use their 80-trial estimates as references and naive as the baseline when $k>m_j$. Appendix~\ref{sec:combinatorial-derivation} gives the formulas.

\subsection{Metric estimation}
\label{sec:mixture-posterior}
We model the observed counts as $Y_j\mid\theta_j\sim\operatorname{Binomial}(m_j,\theta_j)$ and place a Beta mixture prior on the latent success probability, $\theta_j\sim P_\gamma=\sum_{h=1}^{H}w_h\Beta(\alpha_h,\beta_h)$, with positive shapes and weights summing to one. We fit $\gamma$ by marginal maximum likelihood and select $H\in\{1,2,3,4\}$ with the Bayesian information criterion (BIC) \citep{Schwarz1978EstimatingTD}. For brevity, we hereafter refer to the selected Beta mixture fit as \emph{Mixture}. Appendix~\ref{sec:likelihood-derivation} details the model and fitting procedure.

The moments of $\theta^r$ and $(1-\theta)^r$ under each Beta component have closed forms, so the fitted parameters give
\begin{equation*}
\begin{aligned}
 \widehat{\mathrm{Pass}^{k}}_{\rm Mix}
 &=\sum_h\hat w_h\frac{\mathrm B(\hat\alpha_h+k,\hat\beta_h)}{\mathrm B(\hat\alpha_h,\hat\beta_h)},\\
 \widehat{\mathrm{AUC@}k}_{\rm Mix}
 &=1-\frac{1}{k}\sum_{r=1}^{k}\sum_h\hat w_h\frac{\mathrm B(\hat\alpha_h,\hat\beta_h+r)}{\mathrm B(\hat\alpha_h,\hat\beta_h)}.
\end{aligned}
\end{equation*}
One fit gives every horizon, including $k>m_j$.

For a smooth population metric $M$, let $\gamma$ denote the model parameters and $\widehat\Sigma_\gamma$ the inverse observed information. The delta method gives the approximate 95\% interval
\begin{equation*}
 \widehat M\pm1.96
 \sqrt{\nabla M(\hat\gamma)^\top\widehat\Sigma_\gamma\nabla M(\hat\gamma)}.
\end{equation*}
It uses the same fit and requires no extra model calls. The approximation treats the selected $H$ as fixed and relies on the model and regularity conditions in Appendix~\ref{sec:interval-details}. That appendix derives each metric's variance and gradient, including Pass@$k$.

\subsection{Sensitivity analysis}
\label{sec:prompts-method}
For sensitivity analysis, configurations replace tasks as instances. We evaluate this use case with prompts, following PromptEval \citep{polo2024efficient}. With $J$ prompts and $Q$ questions, the complete binary correctness matrix defines
\begin{equation*}
 S_j=\frac{1}{Q}\sum_{q=1}^{Q}Y_{jq},\qquad
 F_S=\frac{1}{J}\sum_{j=1}^{J}\delta_{S_j}.
\end{equation*}
Prompt counts $(y_j,m_j)$ enter the same likelihood and BIC fit. We compare the fitted distribution $\widehat G_{\rm prompt}=P_{\hat\gamma_{\hat H}}$ with $F_S$ using $W_1(\widehat G_{\rm prompt},F_S)$. Appendix~\ref{sec:prompt-derivation} details this and distinguishes latent probabilities from finite-question scores.

%% file: sections/experiments.tex
\section{Experiments and results}
\label{sec:results}

\noindent\textbf{Evaluation protocol.}
We use 104 model--benchmark groups from Scorio \citep{hariri2026test}, with 80 recorded trials per task. Trace and Math contain competition mathematics tasks; GPQA uses 3,600 SuperGPQA questions \citep{du2025supergpqa}. We compare estimates from $m_j=m$ observed trials per task with the 80-trial combinatorial reference. Mean absolute error (MAE) is in percentage points, averaging three subsampling seeds within each group and then groups. Mixture and naive use identical subsets. Appendix~\ref{sec:data} gives details.

\begin{figure}[!ht]
 \centering
 \includegraphics[width=\columnwidth]{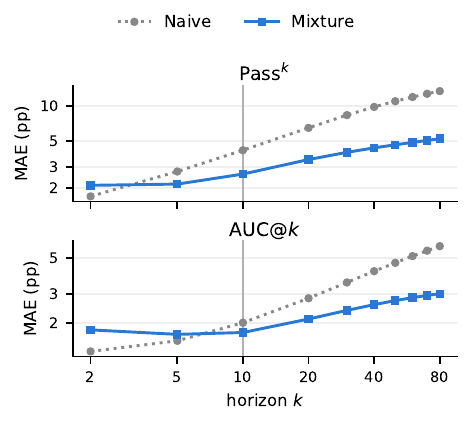}
 \caption{Once $k>m$ (vertical line), naive error grows faster than Mixture's. 104 groups, $m=10$.}
 \label{fig:metrics-k}
\end{figure}

\begin{figure}[!ht]
 \centering
 \includegraphics[width=\columnwidth]{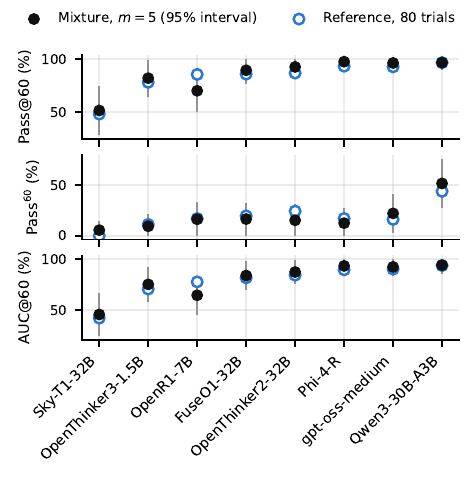}
 \caption{On AIME 2025 (30 tasks, $m=5$), all Mixture intervals cover the reference.}
 \label{fig:models-ci}
\end{figure}

\begin{figure*}[t]
 \centering
 \includegraphics[width=.9\textwidth]{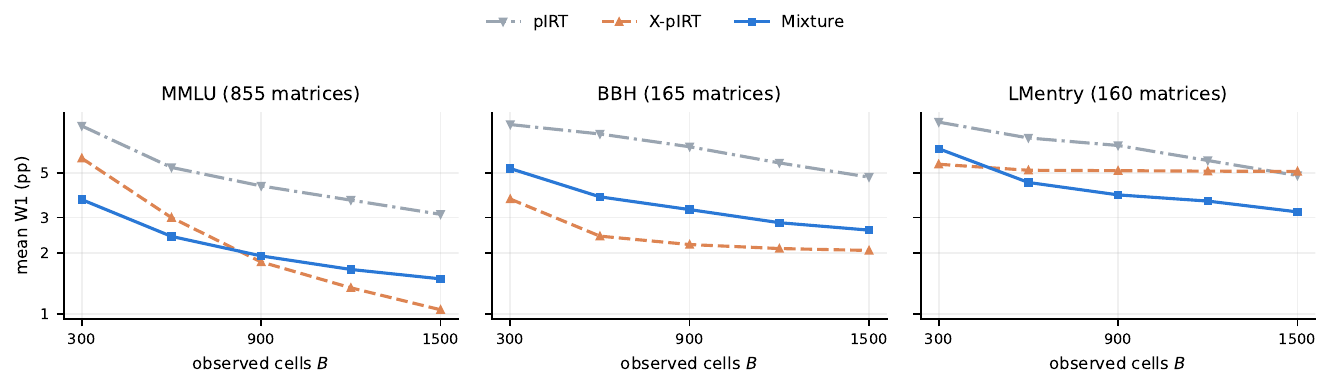}
 \caption{Without external covariates, Mixture beats pIRT at every budget. Mean $W_1$ error of the prompt score distribution by number of observed cells $B$.}
 \label{fig:prompts}
\end{figure*}

\subsection{Repeated success metrics}

\begin{table}[!ht]
 \centering
 \small
 \caption{Mixture cuts error 2--10$\times$ in every dataset. MAE at $k=60$, $m=5$, mean of Pass$^{60}$ and AUC@60.}
 \label{tab:k60-summary}
 \input{tables/k60_summary.tex}
\end{table}

With $m=5$ and $k=60$, Mixture has lower error than naive for most groups (Table~\ref{tab:k60-summary}). GPQA has the largest gain and many more tasks per fit, although models and difficulty also differ. BIC selects one component in every GPQA fit at $m=5$, so this gain does not require a larger mixture.

At $m=10$ and $k=60$, Mixture reduces Pass$^{60}$ MAE from 11.95 to 4.89 points and AUC@60 MAE from 5.13 to 2.86 (Figure~\ref{fig:metrics-k}; Appendix~\ref{sec:dataset-curves} gives curves by dataset). Naive applies a nonlinear transformation to sparse success rates, while Mixture first estimates their shared distribution.

\begin{table}[!ht]
 \centering
 \footnotesize
 \caption{Mixture wins whenever $k>m$; naive wins on Trace and Math when $k\leq m$. MAE by regime.}
 \label{tab:regimes}
 \input{tables/regime.tex}
\end{table}

Mixture has lower mean error in all six collection--metric comparisons for $k>m$ (Table~\ref{tab:regimes}). When $k\leq m$, naive is more accurate on Trace and Math, and combinatorial is more accurate than Mixture in all six comparisons (Appendix~\ref{sec:full-metrics}). At fixed $k=60$, both errors decrease as $m$ grows, but naive becomes more accurate for AUC@60 from about $m=40$. For Pass$^{60}$, Mixture with $m=10$ matches naive with roughly $m=30$. The main benefit is therefore extrapolation from sparse counts. In practice, we recommend combinatorial when $m_j\geq k$ and Mixture for larger horizons and for intervals. Appendices~\ref{sec:full-metrics} and~\ref{sec:seed-variation} give budget curves and seed variability.

\noindent\textbf{Confidence intervals.} Figure~\ref{fig:models-ci} shows the intervals on AIME 2025 with $m=5$ and $k=60$ for eight models evenly spaced by reference AUC@60 (all 20 in Appendix~\ref{sec:models-ci}). For all 20 models and all three metrics, the 95\% interval contains the 80-trial reference. With 30 tasks and five trials, the intervals are wide (25 to 30 points), because they include uncertainty from the finite task set.

\subsection{Sensitivity across prompts}
We use PromptEval matrices for MMLU, BBH, and LMentry with three sampling seeds \citep{polo2024efficient}. All methods observe the same cells. Mixture has lower mean $W_1$ than pIRT at every tested budget on all three benchmarks, with a 46\% mean relative reduction across the 15 benchmark--budget comparisons (Figure~\ref{fig:prompts}). At 300 observed cells on MMLU, errors are 3.68 and 8.53 points. After seed averaging, Mixture wins on 93\% of MMLU matrices, 88\% of BBH matrices, and 81\% of LMentry matrices at this budget.

X-pIRT is more accurate on BBH and at larger MMLU budgets, but uses external covariates such as embeddings. Mixture and pIRT use only observed outcomes. Appendix~\ref{sec:prompt-details} reports all budgets, distribution curves, and quantile errors.

%% file: tables/k60_summary.tex
\begin{tabular*}{\columnwidth}{@{\extracolsep{\fill}}lrrrr}
\toprule
Dataset & Tasks & Naive & Mixture & Win rate (\%)\\
\midrule
Trace & 30 & 12.90 & \textbf{4.84} & 96 \\
Math & 30--53 & 14.77 & \textbf{6.55} & 95 \\
GPQA & 3,600 & 13.65 & \textbf{1.42} & 100 \\
\bottomrule
\end{tabular*}

%% file: tables/regime.tex
\setlength{\tabcolsep}{3pt}
\begin{tabular*}{\columnwidth}{@{\extracolsep{\fill}}llrrrr}
\toprule
& & \multicolumn{2}{c}{$k\leq m$} & \multicolumn{2}{c}{$k>m$}\\
Dataset & Metric & Naive & Mixture & Naive & Mixture\\
\midrule
Trace & Pass$^k$ & \textbf{1.58} & 1.85 & 7.07 & \textbf{3.51} \\
Trace & AUC@$k$ & \textbf{0.83} & 1.46 & 3.24 & \textbf{2.21} \\
\addlinespace
Math & Pass$^k$ & \textbf{1.93} & 2.34 & 9.46 & \textbf{4.97} \\
Math & AUC@$k$ & \textbf{0.66} & 1.38 & 2.38 & \textbf{2.06} \\
\addlinespace
GPQA & Pass$^k$ & 1.19 & \textbf{0.16} & 5.95 & \textbf{0.83} \\
GPQA & AUC@$k$ & 0.83 & \textbf{0.16} & 4.64 & \textbf{0.38} \\
\bottomrule
\end{tabular*}

%% file: sections/conclusion.tex
\section{Conclusion}
A fitted Beta mixture supports repeated success estimation and sensitivity analysis from sparse outcomes. It reduces error relative to naive for Pass$^k$ and AUC@$k$ when $k>m$, while direct estimators can be more accurate when observations reach the target horizon. Treating prompts as instances estimates sensitivity to prompt choice with lower distribution error than comparable baselines.

\section*{Limitations}
The model assumes conditional independence and does not account for shared question difficulty. The 80-trial reference is a finite-sample estimate, and the task data are limited to reasoning benchmarks. The sensitivity experiments vary only the prompts. Furthermore, BIC may not be the most appropriate criterion for selecting the mixture model parameters.

%% file: sections/appendix.tex
\appendix
\raggedbottom
\onecolumn

\section{Metric estimators and interval derivations}
\label{sec:derivations}
We derive the marginal likelihood, the moments used by the metrics, and their delta method intervals. Throughout, $m_j$ is the number of observed attempts on instance $j$, $y_j$ is its success count, $k$ is the target horizon, and $H$ is the number of mixture components. The population target is an expectation over $\theta\sim P_\gamma$. Its finite-benchmark counterpart averages the same function over the probabilities of the evaluated instances. These two targets need not coincide for a fixed benchmark.

\subsection{Beta function identities}
\label{sec:beta-facts}
For $\alpha,\beta>0$, the Beta function and density are
\[
 \mathrm B(\alpha,\beta)=\int_0^1 t^{\alpha-1}(1-t)^{\beta-1}\,dt
 =\frac{\Gamma(\alpha)\Gamma(\beta)}{\Gamma(\alpha+\beta)},
 \qquad
 \Beta(\theta;\alpha,\beta)=
 \frac{\theta^{\alpha-1}(1-\theta)^{\beta-1}}{\mathrm B(\alpha,\beta)}.
\]
We use $\Gamma(x+1)=x\Gamma(x)$ and the digamma function $\psi(x)=d\log\Gamma(x)/dx$. For an integer $r\geq0$,
\[
 \psi(x+r)-\psi(x)=\sum_{i=0}^{r-1}\frac{1}{x+i}.
\]
An empty sum is zero, and an empty product is one.

\subsection{Beta mixture and maximum likelihood fit}
\label{sec:likelihood-derivation}
For a fixed system and evaluation setting, instances are independent draws from the shared distribution,
\begin{equation*}
\begin{aligned}
 \theta_j&\sim P_\gamma=\sum_{h=1}^{H}w_h\Beta(\alpha_h,\beta_h),\\
 Y_j\mid\theta_j,m_j&\sim\operatorname{Binomial}(m_j,\theta_j).
\end{aligned}
\end{equation*}
Here $\gamma=\{w_h,\alpha_h,\beta_h\}_{h=1}^{H}$, with $w_h>0$, $\sum_h w_h=1$, and $\alpha_h,\beta_h>0$. Within an instance, attempts are independent conditional on $\theta_j$. For component $h$, integration gives
\begin{align*}
 \mathbb P_h(Y_j=y_j\mid m_j)
 &=\frac{\binom{m_j}{y_j}}{\mathrm B(\alpha_h,\beta_h)}
 \int_0^1\theta^{\alpha_h+y_j-1}(1-\theta)^{\beta_h+m_j-y_j-1}\,d\theta\\
 &=\binom{m_j}{y_j}
 \frac{\mathrm B(\alpha_h+y_j,\beta_h+m_j-y_j)}
 {\mathrm B(\alpha_h,\beta_h)}.
\end{align*}
Summing over components yields the marginal probability,
\begin{equation*}
 \mathbb P_\gamma(Y_j=y_j\mid m_j)=
 \binom{m_j}{y_j}\sum_{h=1}^{H}w_h
 \frac{\mathrm B(\alpha_h+y_j,\beta_h+m_j-y_j)}
 {\mathrm B(\alpha_h,\beta_h)}.
\end{equation*}
At fixed $H$, the marginal maximum likelihood estimate is
\begin{equation*}
 \hat\gamma_H=\arg\max_{\gamma}\ell(\gamma),
 \qquad
 \ell(\gamma)=\sum_{j:m_j>0}\log\mathbb P_\gamma(Y_j=y_j\mid m_j).
\end{equation*}
The maximum is over all mixtures with $H$ components. An instance with $m_j=0$ contributes probability one and contains no information about $\gamma$. Let $J_{\rm obs}$ count the instances with positive observed budgets. The parameter count is $3H-1$, comprising $H$ pairs of shapes and $H-1$ free weights. We select
\begin{equation*}
\begin{aligned}
 \hat H&=\arg\min_{H\in\{1,2,3,4\}}\operatorname{BIC}(H),\\
 \operatorname{BIC}(H)&=-2\ell(\hat\gamma_H)+(3H-1)\log J_{\rm obs}.
\end{aligned}
\end{equation*}
This is the Bayesian information criterion \citep{Schwarz1978EstimatingTD}. In the repeated-task experiment, $J_{\rm obs}=J$. For prompts, only prompts with observed answers enter the fit and the BIC sample count. The reference distribution still includes every evaluated prompt.

The implementation uses L-BFGS-B with ten initializations per candidate. For $H>1$, it also initializes from a split of the previous candidate's fit. The solution with the largest likelihood is retained. Hats below denote parameters of the selected fit $\hat\gamma_{\hat H}$.

\subsection{Beta moments}
\label{sec:beta-moments}
\begin{lemma}
\label{lem:beta-moments}
If $\theta\sim\Beta(\alpha,\beta)$ and $r\geq0$ is an integer, then
\[
 \E[\theta^r]=\frac{\mathrm B(\alpha+r,\beta)}{\mathrm B(\alpha,\beta)}
 =\prod_{i=0}^{r-1}\frac{\alpha+i}{\alpha+\beta+i},
 \qquad
 \E[(1-\theta)^r]=\frac{\mathrm B(\alpha,\beta+r)}{\mathrm B(\alpha,\beta)}
 =\prod_{i=0}^{r-1}\frac{\beta+i}{\alpha+\beta+i}.
\]
\end{lemma}
\begin{proof}
Multiplying the density by $\theta^r$ and integrating gives
\[
 \E[\theta^r]=
 \frac{1}{\mathrm B(\alpha,\beta)}
 \int_0^1\theta^{\alpha+r-1}(1-\theta)^{\beta-1}\,d\theta
 =\frac{\mathrm B(\alpha+r,\beta)}{\mathrm B(\alpha,\beta)}.
\]
Writing the ratio with Gamma functions and applying $\Gamma(x+1)=x\Gamma(x)$ gives the product. The second identity follows by exchanging $\alpha$ and $\beta$, since $1-\theta\sim\Beta(\beta,\alpha)$.
\end{proof}
For a mixture, $\E_\gamma[f(\theta)]=\sum_h w_h\E_h[f(\theta)]$, where $\E_h$ is the expectation under $\Beta(\alpha_h,\beta_h)$. Each metric below is a weighted sum of these component moments.

\subsection{Delta method intervals}
\label{sec:interval-details}
\paragraph{Parameters and covariance.}
For fixed $H$, the parameters $\gamma=\{w_h,\alpha_h,\beta_h\}_{h=1}^{H}$ of Appendix~\ref{sec:likelihood-derivation} have $3H-1$ free entries, the number counted by BIC, because the weights sum to one. We write $w_1=1-\sum_{h\geq2}w_h$, so gradients and $\widehat\Sigma_\gamma$ below are taken with respect to $w_2,\ldots,w_H$ and all shapes. Under a correctly specified independent-instance model, local identifiability, an interior optimum, and nonsingular information, the maximum likelihood estimate $\hat\gamma$ is asymptotically normal as the number of instances increases. Its covariance is estimated by the inverse observed information
\[
 \widehat\Sigma_\gamma=
 \left[-\nabla_\gamma^2\ell(\gamma)\big|_{\hat\gamma}\right]^{-1}.
\]

\paragraph{Metric variance.}
For a smooth metric $M(\gamma)$, a first-order Taylor expansion gives
\[
 M(\hat\gamma)\approx M(\gamma)+\nabla_\gamma M(\gamma)^\top(\hat\gamma-\gamma).
\]
With $\hat g_M=\nabla_\gamma M(\hat\gamma)$,
\begin{equation}
 \widehat V_M=\hat g_M^\top\widehat\Sigma_\gamma\hat g_M,
 \qquad
 \mathcal I_M=\left[\widehat M-1.96\sqrt{\widehat V_M},
                    \widehat M+1.96\sqrt{\widehat V_M}\right].
\label{eq:delta}
\end{equation}
The full covariance matrix is used, including covariance between components.

\paragraph{Component derivatives.}
Each metric below is a weighted sum over components. Since $w_1=1-\sum_{h\geq2}w_h$, the derivative with respect to $w_h$, $h\geq2$, is the term of component $h$ minus the term of component 1. Differentiating the log Beta ratios gives, for an integer $r\geq0$,
\begin{align*}
 \partial_{\alpha}\frac{\mathrm B(\alpha+r,\beta)}{\mathrm B(\alpha,\beta)}
 &=\frac{\mathrm B(\alpha+r,\beta)}{\mathrm B(\alpha,\beta)}\bigl[\psi(\alpha+r)-\psi(\alpha)
               +\psi(\alpha+\beta)-\psi(\alpha+\beta+r)\bigr],\\
 \partial_{\beta}\frac{\mathrm B(\alpha+r,\beta)}{\mathrm B(\alpha,\beta)}
 &=\frac{\mathrm B(\alpha+r,\beta)}{\mathrm B(\alpha,\beta)}\bigl[\psi(\alpha+\beta)-\psi(\alpha+\beta+r)\bigr],\\
 \partial_{\alpha}\frac{\mathrm B(\alpha,\beta+r)}{\mathrm B(\alpha,\beta)}
 &=\frac{\mathrm B(\alpha,\beta+r)}{\mathrm B(\alpha,\beta)}\bigl[\psi(\alpha+\beta)-\psi(\alpha+\beta+r)\bigr],\\
 \partial_{\beta}\frac{\mathrm B(\alpha,\beta+r)}{\mathrm B(\alpha,\beta)}
 &=\frac{\mathrm B(\alpha,\beta+r)}{\mathrm B(\alpha,\beta)}\bigl[\psi(\beta+r)-\psi(\beta)
               +\psi(\alpha+\beta)-\psi(\alpha+\beta+r)\bigr].
\end{align*}
Below, these derivatives are evaluated at $(\alpha_h,\beta_h)$, and $\nabla_h=(\partial_{\alpha_h},\partial_{\beta_h})^\top$.

\paragraph{Implementation.}
The optimizer works in unconstrained coordinates. Weights are a softmax of $H-1$ free logits, component means are ordered on the logit scale, and concentrations are on the log scale. It also uses box bounds on these coordinates to avoid overflow, and an optimum on these bounds does not meet the interior condition above. We compute the Hessian and $\hat g_M$ in these coordinates by central differences, and use the pseudo-inverse if the Hessian is singular. At an interior optimum, this gives the same interval as Equation~\ref{eq:delta}, because the delta method interval does not change under a smooth one-to-one change of parameters.

\paragraph{Scope.}
The approximation treats the selected $H$ as fixed and omits selection uncertainty. Very small weights or nearly coincident components can make the observed information ill-conditioned.

\subsection{Pass\texorpdfstring{$^k$}{\^{}k}}
\label{sec:passhat-derivation}
Conditional independence gives probability $\theta^k$ of success in every attempt. Lemma~\ref{lem:beta-moments} then gives
\[
 \mathrm{Pass}^{k}(\gamma)=\sum_h w_h\frac{\mathrm B(\alpha_h+k,\beta_h)}{\mathrm B(\alpha_h,\beta_h)}.
\]
The estimate is $\widehat{\mathrm{Pass}^{k}}_{\rm Mix}=\mathrm{Pass}^{k}(\hat\gamma)$, as in Section~\ref{sec:mixture-posterior}. The derivatives are
\[
 \begin{pmatrix}g_{\alpha_h}\\g_{\beta_h}\end{pmatrix}_{\rm Pass^k}=w_h\nabla_h\frac{\mathrm B(\alpha_h+k,\beta_h)}{\mathrm B(\alpha_h,\beta_h)},
 \qquad
 g_{w_h,\rm Pass^k}=\frac{\mathrm B(\alpha_h+k,\beta_h)}{\mathrm B(\alpha_h,\beta_h)}-\frac{\mathrm B(\alpha_1+k,\beta_1)}{\mathrm B(\alpha_1,\beta_1)}.
\]
The variance is
$\widehat V_{\rm Pass^k}=\hat g_{\rm Pass^k}^\top\widehat\Sigma_\gamma\hat g_{\rm Pass^k}$,
and $\widehat{\mathrm{Pass}^{k}}_{\rm Mix}\pm1.96\sqrt{\widehat V_{\rm Pass^k}}$.

\subsection{Pass@\texorpdfstring{$k$}{k}}
\label{sec:pass-derivation}
At least one attempt succeeds unless all attempts fail. Thus
\[
 \mathrm{Pass@}k(\gamma)=1-\sum_h w_h\frac{\mathrm B(\alpha_h,\beta_h+k)}{\mathrm B(\alpha_h,\beta_h)}.
\]
The estimate is $\widehat{\mathrm{Pass@}k}_{\rm Mix}=\mathrm{Pass@}k(\hat\gamma)$. The shape derivatives have the opposite sign to those of the probability that all $k$ attempts fail,
\[
 \begin{pmatrix}g_{\alpha_h}\\g_{\beta_h}\end{pmatrix}_{\rm Pass@k}=-w_h\nabla_h\frac{\mathrm B(\alpha_h,\beta_h+k)}{\mathrm B(\alpha_h,\beta_h)},
 \qquad
 g_{w_h,\rm Pass@k}=\frac{\mathrm B(\alpha_1,\beta_1+k)}{\mathrm B(\alpha_1,\beta_1)}-\frac{\mathrm B(\alpha_h,\beta_h+k)}{\mathrm B(\alpha_h,\beta_h)}.
\]
The variance and interval are
$\widehat V_{\rm Pass@k}=\hat g_{\rm Pass@k}^\top\widehat\Sigma_\gamma\hat g_{\rm Pass@k}$
and $\widehat{\mathrm{Pass@}k}_{\rm Mix}\pm1.96\sqrt{\widehat V_{\rm Pass@k}}$.

\subsection{AUC@\texorpdfstring{$k$}{k}}
\label{sec:auc-derivation}
Averaging Pass@$r$ over $r=1,\ldots,k$ gives
\[
 \mathrm{AUC@}k(\gamma)=1-\frac{1}{k}\sum_{r=1}^{k}\sum_h w_h\frac{\mathrm B(\alpha_h,\beta_h+r)}{\mathrm B(\alpha_h,\beta_h)}.
\]
and $\widehat{\mathrm{AUC@}k}_{\rm Mix}=\mathrm{AUC@}k(\hat\gamma)$. The same averaging applies to the derivatives,
\[
\begin{aligned}
 \begin{pmatrix}g_{\alpha_h}\\g_{\beta_h}\end{pmatrix}_{\rm AUC@k}
 &=-\frac{w_h}{k}\sum_{r=1}^{k}\nabla_h\frac{\mathrm B(\alpha_h,\beta_h+r)}{\mathrm B(\alpha_h,\beta_h)},\\
 g_{w_h,\rm AUC@k}
 &=\frac{1}{k}\sum_{r=1}^{k}\left[\frac{\mathrm B(\alpha_1,\beta_1+r)}{\mathrm B(\alpha_1,\beta_1)}-\frac{\mathrm B(\alpha_h,\beta_h+r)}{\mathrm B(\alpha_h,\beta_h)}\right].
\end{aligned}
\]
In particular, $\hat g_{\rm AUC@k}=k^{-1}\sum_{r=1}^{k}\hat g_{\rm Pass@r}$. Hence
\[
 \widehat V_{\rm AUC@k}
 =\frac{1}{k^2}\sum_{r=1}^{k}\sum_{s=1}^{k}
 \hat g_{\rm Pass@r}^{\top}\widehat\Sigma_\gamma\hat g_{\rm Pass@s}.
\]
This includes the covariance between horizons, which come from the same fit. The interval is
$\widehat{\mathrm{AUC@}k}_{\rm Mix}\pm1.96\sqrt{\widehat V_{\rm AUC@k}}$.
The recursion $\mathrm B(\alpha_h,\beta_h+r+1)=\mathrm B(\alpha_h,\beta_h+r)(\beta_h+r)/(\alpha_h+\beta_h+r)$ computes the terms in $O(Hk)$ operations.

\subsection{One-component formulas}
\label{sec:one-component}
BIC selects $H=1$ in 95\% of the repeated-task fits (Appendix~\ref{sec:k-selection}). In that case the Beta ratios become finite products.
\begin{proposition}
\label{prop:one-component}
For $\theta\sim\Beta(\alpha,\beta)$ and $k\geq1$,
\[
 \mathrm{Pass}^{k}=\prod_{i=0}^{k-1}\frac{\alpha+i}{\alpha+\beta+i},\qquad
 1-\mathrm{Pass@}k=\prod_{i=0}^{k-1}\frac{\beta+i}{\alpha+\beta+i}.
\]
The log derivatives of Pass$^k$ are
\[
 \partial_\alpha\log\mathrm{Pass}^{k}
 =\sum_{i=0}^{k-1}\left[\frac{1}{\alpha+i}-\frac{1}{\alpha+\beta+i}\right],
 \qquad
 \partial_\beta\log\mathrm{Pass}^{k}
 =-\sum_{i=0}^{k-1}\frac{1}{\alpha+\beta+i}.
\]
Exchanging $\alpha$ and $\beta$ gives the corresponding derivatives of $\log(1-\mathrm{Pass@}k)$.
\end{proposition}
\begin{proof}
The products follow from Lemma~\ref{lem:beta-moments}. Taking logs and differentiating each factor gives the derivative sums.
\end{proof}
For fixed positive shapes, as $k\to\infty$,
\[
 \mathrm{Pass}^{k}\sim\frac{\Gamma(\alpha+\beta)}{\Gamma(\alpha)}k^{-\beta},
 \qquad
 1-\mathrm{Pass@}k\sim\frac{\Gamma(\alpha+\beta)}{\Gamma(\beta)}k^{-\alpha}.
\]
These follow from $\Gamma(k+x)/\Gamma(k+z)\sim k^{x-z}$. The exponents depend on the behavior of the Beta density near one and zero, respectively.

\subsection{Combinatorial and naive estimators}
\label{sec:combinatorial-derivation}
When $k\leq m_j$ for every instance, the unbiased benchmark estimates are
\begin{equation}
\begin{aligned}
 \widehat{\mathrm{Pass@}k}_{\rm comb}
 &=\frac{1}{J}\sum_j\left[1-\frac{\binom{m_j-y_j}{k}}{\binom{m_j}{k}}\right],\\
 \widehat{\mathrm{Pass}^{k}}_{\rm comb}
 &=\frac{1}{J}\sum_j\frac{\binom{y_j}{k}}{\binom{m_j}{k}}.
\end{aligned}
\label{eq:combinatorial}
\end{equation}
A numerator is zero if fewer than $k$ failures or successes are available. For $Y_j\sim\operatorname{Binomial}(m_j,\theta_j)$, counting subsets of successful or failed attempts gives
\[
 \E\left[\binom{Y_j}{k}\mid\theta_j\right]=\binom{m_j}{k}\theta_j^k,\qquad
 \E\left[\binom{m_j-Y_j}{k}\mid\theta_j\right]=\binom{m_j}{k}(1-\theta_j)^k.
\]
Dividing by $\binom{m_j}{k}$ and averaging proves unbiasedness for the benchmark metrics given the $\theta_j$. Averaging the Pass@$r$ estimates for $r=1,\ldots,k$ also gives an unbiased AUC@$k$ estimate. None of these estimators is defined if some $m_j<k$.

For naive, $\hat p_j=y_j/m_j$ has conditional mean $\theta_j$. When $k>1$, convexity of $p^k$ and concavity of $1-(1-p)^k$ give upward Pass$^k$ bias and downward Pass@$k$ bias by Jensen's inequality. Averaging the latter over horizons gives downward AUC@$k$ bias.

Let $y_j^{(80)}$ be the success count of instance $j$ in all 80 recorded trials. The AUC reference is
\[
 \widehat{\mathrm{AUC@}k}_{80}
 =\frac{1}{kJ}\sum_{r=1}^{k}\sum_{j=1}^{J}
 \left[1-\frac{\binom{80-y_j^{(80)}}{r}}{\binom{80}{r}}\right].
\]
It averages every integer horizon up to $k$, not only the plotted ones.

\subsection{Sensitivity analysis with prompts}
\label{sec:prompt-derivation}
Prompt $j$ is an instance, with $y_j$ observed correct answers among $m_j$ sampled questions. We use the same binomial model and marginal likelihood as above. The observed cells determine the fit, and the full outcome matrix gives only the reference scores. The model assumes that the answers of one prompt are independent with a common success probability. This is an approximation, since questions differ in difficulty.

\paragraph{Fitted distribution and quantiles.}
The fitted distribution function is
\[
 \widehat G_{\rm prompt}(x)=\sum_{h=1}^{\hat H}\hat w_h
 I_x(\hat\alpha_h,\hat\beta_h),
\]
where $I_x(\alpha,\beta)$ is the regularized incomplete Beta function. We compute its inverse with 60 bisection steps on $[0,1]$.

\paragraph{Distributional comparison.}
For distributions on the real line,
\[
 W_1(G,F)=\int_0^1\left|G^{-1}(u)-F^{-1}(u)\right|\,du.
\]
Let $S_{(1)}\leq\cdots\leq S_{(J)}$ be the sorted complete prompt scores. Evaluating the fitted quantiles at $u_i=(i-\tfrac12)/J$ gives the approximation used in the experiments,
\[
 W_1(\widehat G_{\rm prompt},F_S)\approx
 \frac{1}{J}\sum_{i=1}^{J}
 \left|\widehat G_{\rm prompt}^{-1}(u_i)-S_{(i)}\right|.
\]
We also compare the 5th, 25th, 50th, 75th, and 95th percentiles of these $J$ fitted quantile values with the corresponding percentiles of the reference scores, following PromptEval \citep{polo2024efficient}.

\paragraph{Latent probabilities and finite scores.}
Under the working model of $Q$ conditionally independent answers with probability $\theta_j$,
$\E[S_j\mid\theta_j]=\theta_j$ and $\Var(S_j\mid\theta_j)=\theta_j(1-\theta_j)/Q$. By the law of total variance,
\[
 \E[S_j]=\E[\theta_j],\qquad
 \Var(S_j)=\Var(\theta_j)+\frac{\E[\theta_j(1-\theta_j)]}{Q}.
\]
The latent distribution and the distribution of complete scores have the same mean but different variances at finite $Q$. Shared question difficulty can change this relation further. The experiment therefore measures agreement with the finite prompt scores, not recovery of the latent distribution.

\twocolumn
\setlength{\emergencystretch}{1.5em}
\renewcommand{\topfraction}{.85}
\renewcommand{\dbltopfraction}{.85}
\renewcommand{\textfraction}{.1}
\renewcommand{\floatpagefraction}{.6}
\renewcommand{\dblfloatpagefraction}{.6}
\makeatletter
\setlength{\@fptop}{0pt}
\setlength{\@dblfptop}{0pt}
\setlength{\@fpsep}{16pt}
\setlength{\@dblfpsep}{16pt}
\makeatother
\setcounter{topnumber}{3}
\setcounter{dbltopnumber}{3}
\setcounter{totalnumber}{4}

\section{Datasets and experimental design}
\label{sec:data}

\paragraph{Repeated-task datasets.}
We use three collections of repeated task outcomes from the Scorio datasets \citep{hariri2026test}, which we call Trace, Math, and GPQA. Trace contains AIME 2024, AIME 2025, BrUMO 2025, and HMMT February 2025, with 30 tasks in each benchmark. Math contains AIME 2026 with 30 tasks, CMIMC 2025 with 40, HMMT February 2026 with 33, HMMT November 2025 with 30, and SMT 2025 with 53. GPQA contains a fixed sample of 3,600 questions from SuperGPQA \citep{du2025supergpqa}, with 50 questions from each of 72 fields. Every task has 80 recorded trials. A group is one model evaluated on one benchmark, giving 80 groups for Trace, 20 for Math, and 4 for GPQA.

\paragraph{Models.}
Trace evaluates 20 open models. Math and GPQA evaluate Qwen3.6-35B-A3B and \texttt{gpt-oss-20b} with low, medium, and high reasoning. Table~\ref{tab:models} lists every model with the short name used in figures.

\begin{table*}[t]
 \centering
 \small
 \caption{Models, short names, and the collections that use them.}
 \label{tab:models}
 \input{tables/models.tex}
\end{table*}

\paragraph{Repeated-task experimental design.}
For each of three random seeds, we independently permute the 80 trials of every task and use the first $m$ trials as a nested subset, with $m\in\{5,10,15,20,\allowbreak 25,30,35,40,\allowbreak 50,60,70,80\}$. We estimate metrics at $k\in\{2,5,10,20,30,\allowbreak 40,50,60,70,80\}$. All methods use the same subset for a given group, $m$, and seed. The reference applies the combinatorial estimators of Appendix~\ref{sec:combinatorial-derivation} to all 80 trials. It is an estimate of each group's benchmark metric, not the population value. We report MAE from this reference in percentage points, first averaging the three seeds within each group and then averaging groups. Appendix~\ref{sec:seed-variation} reports the variation across seeds. We also report results by dataset, because the three collections differ in the number of groups and in the number of tasks per group.

\paragraph{Prompt evaluation.}
We use the PromptEval outcome matrices \citep{polo2024efficient} for MMLU \citep{hendrycks2021measuring}, BBH \citep{suzgun2023challenging}, and LMentry \citep{efrat2023lmentry}. The BBH and LMentry matrices come from \citet{mizrahi2024state}. MMLU contains 855 matrices with 100 prompts each, BBH contains 165 matrices with 136 to 189 prompts, and LMentry contains 160 matrices with 226 to 265 prompts. Each matrix has 26 to 1,534 benchmark questions, depending on the dataset and matrix. We observe $B\in\{300,600,900,1200,1500\}$ cells (prompt and question pairs) with PromptEval's stratified sampling and use three seeds. The complete matrix determines the reference prompt-score distribution. We report $W_1$ in percentage points of accuracy. We compare with pIRT, which fits prompt and question effects from the outcomes alone, and X-pIRT, which also uses prompt and question covariates such as embeddings \citep{polo2024efficient}. All three methods see the same cells. Our main comparison is with pIRT, because neither method uses external features.

\section{Computing resources}
\label{sec:computing}
All fits run on CPU, on one machine with an Intel Xeon w3-2435 (8 cores) and 125 GiB of RAM. One BIC fit ($H=1$ to 4, ten initializations each) takes at most 1.4 seconds on one core. The naive estimator takes microseconds, but a fit still costs far less than generating a single trial. The repeated-task experiment takes about 25 minutes of single-core time, and the prompt experiment about 2 hours. We did not run any LLM, since all outcomes come from released datasets.

\section{Example in the introduction}
\label{sec:teaser-details}
Figure~\ref{fig:teaser} shows the GPQA group of \texttt{gpt-oss-20b} with medium reasoning, with $m=5$ trials per task and sampling seed 0. We chose a GPQA group because its 3,600 tasks give narrow intervals, which makes the figure easy to read. Seed 0 has the narrowest intervals of the three seeds. BIC selects one component. Over the plotted horizons, the mean interval width is 2.67 points and the largest error is 1.77 points. This is one group, and Section~\ref{sec:results} reports errors over all groups.

\section{Confidence intervals for all models}
\label{sec:models-ci}
\begin{figure}[tb]
 \centering
 \includegraphics[width=\columnwidth]{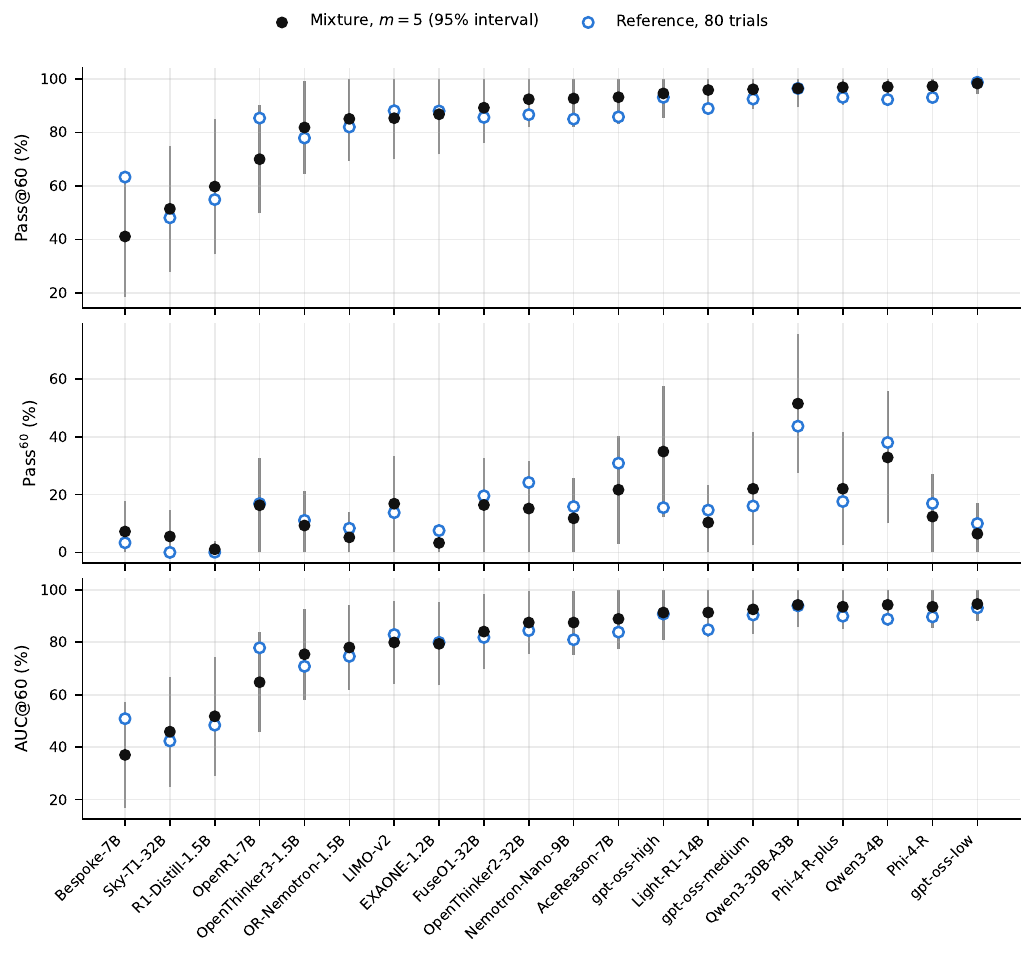}
 \caption{All 60 intervals cover the 80-trial reference. Same as Figure~\ref{fig:models-ci} for all 20 Trace models.}
 \label{fig:models-ci-all}
\end{figure}

Figure~\ref{fig:models-ci-all} extends Figure~\ref{fig:models-ci} to the 20 Trace models on AIME 2025, ordered by the estimated Pass@60. All estimates use sampling seed 0, $m=5$ trials per task, and $k=60$. BIC selects one component for every model. For every model and metric, the 95\% interval contains the 80-trial reference. The mean interval width is 25 points for Pass@60, 30 points for Pass$^{60}$, and 27 points for AUC@60. Near 0 and 100, we clip the interval to the range of the metric.

\section{Detailed repeated success results}
\label{sec:full-metrics}
\begin{figure*}[t]
 \centering
 \includegraphics[width=.8\textwidth]{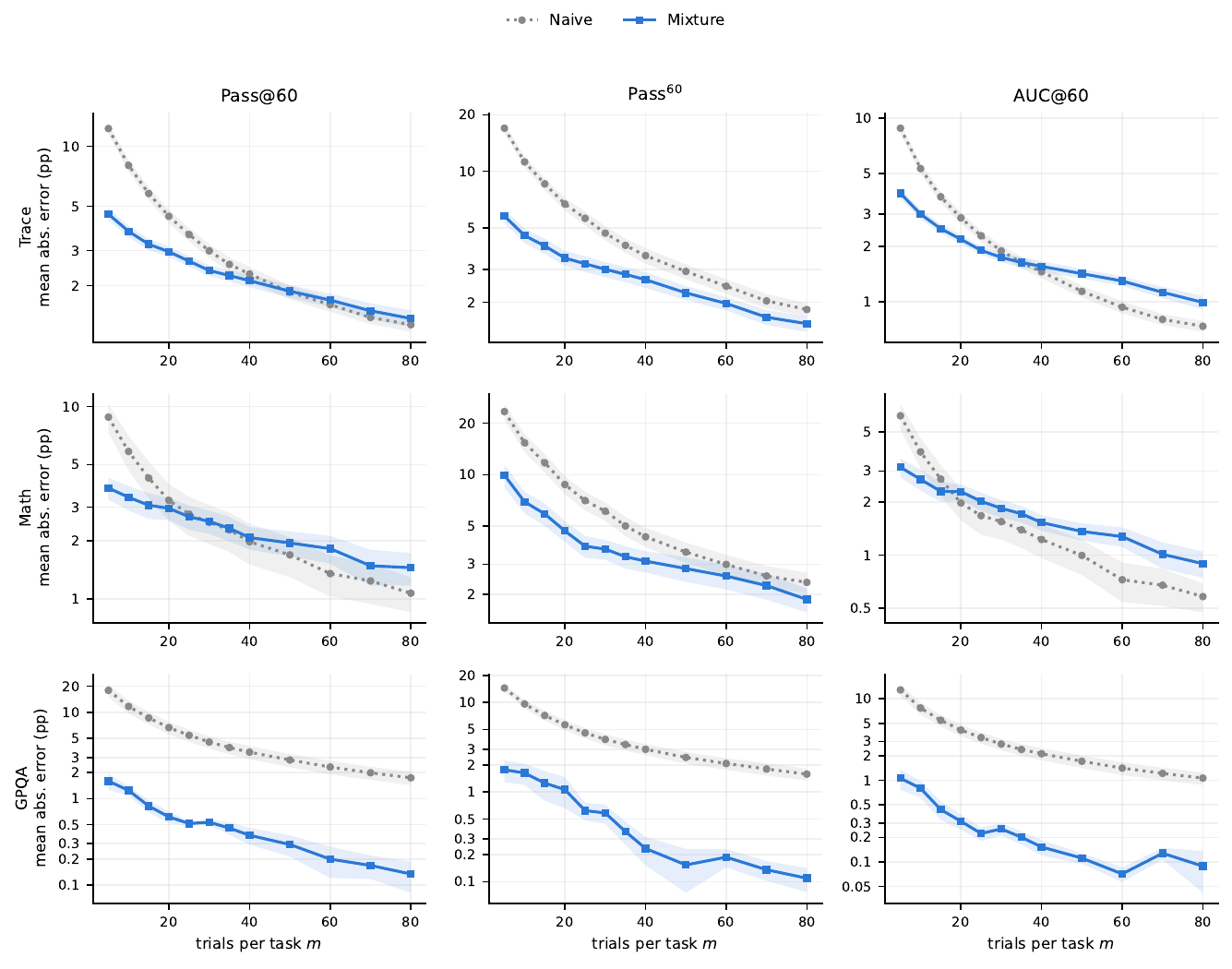}
 \caption{Both errors fall as $m$ grows; on Trace and Math, naive overtakes Mixture for Pass@60 and AUC@60 at large $m$. MAE at $k=60$.}
 \label{fig:full-m}
\end{figure*}

Table~\ref{tab:regimes-full} adds the combinatorial estimator to Table~\ref{tab:regimes}. The tables and figures in this appendix also include Pass@$k$. Tables~\ref{tab:k60-trace}--\ref{tab:k60-gpqa} fix $k=60$ and vary $m$. The combinatorial estimator needs $m\geq60$, and at $m=80$ its error is zero because it equals the reference. Figure~\ref{fig:full-m} shows the same comparison as curves for each dataset and metric. On Trace and Math, naive becomes more accurate than Mixture for Pass@60 and AUC@60 at large $m$. On GPQA, Mixture is more accurate at every $m$.

\paragraph{Trials needed for a given error.}
We compute a retrospective trial-budget comparison at $k=60$. For each group, we average absolute error over Pass$^{60}$, AUC@60, and the three seeds. We then take the smallest tested $m$ whose error is at most 3 points. Cost is proportional to $m$, since every task receives $m$ trials, so the saving relative to the 80 trials is $1-m/80$. A group that never reaches 3 points counts as $m=80$. The median over the 104 groups is $m=15$ for Mixture, which gives 81\% fewer trials ($80/15\approx5.3\times$). The median for naive is $m=35$. Mixture reaches this tolerance in 96\% of the groups. This calculation uses the full-data reference retrospectively and is not a stopping rule available during evaluation.
Table~\ref{tab:budget} repeats this for tolerances from 2 to 7 points and for each collection. Fewer is $1-m/80$ for Mixture. The smallest tested value is $m=5$, so a value of 5 means the tolerance is met at the smallest budget, and 94\% is the largest saving we can measure.

\begin{table*}[t]
 \centering
 \footnotesize
 \setlength{\tabcolsep}{4pt}
 \caption{Mixture reaches every tolerance with fewer trials than naive; at 3 points it needs 15 trials per task instead of 35. Median trials per task needed to reach each error tolerance at $k=60$.}
 \label{tab:budget}
 \input{tables/budget.tex}
\end{table*}

\section{Variation across sampling seeds}
\label{sec:seed-variation}
Table~\ref{tab:seed-variation} reports the error for $k>m$ for each of the three sampling seeds. For each seed, we average the absolute error over the tested $(m,k)$ pairs with $k>m$ within each group, and then over the groups of a collection. The last columns give the mean and the sample standard deviation over seeds. This standard deviation measures sensitivity to the random subset of trials. It is not the same as the delta method interval.

\section{Prompt distribution results}
\label{sec:prompt-details}
\begin{figure*}[p]
 \centering
 \includegraphics[width=.95\textwidth]{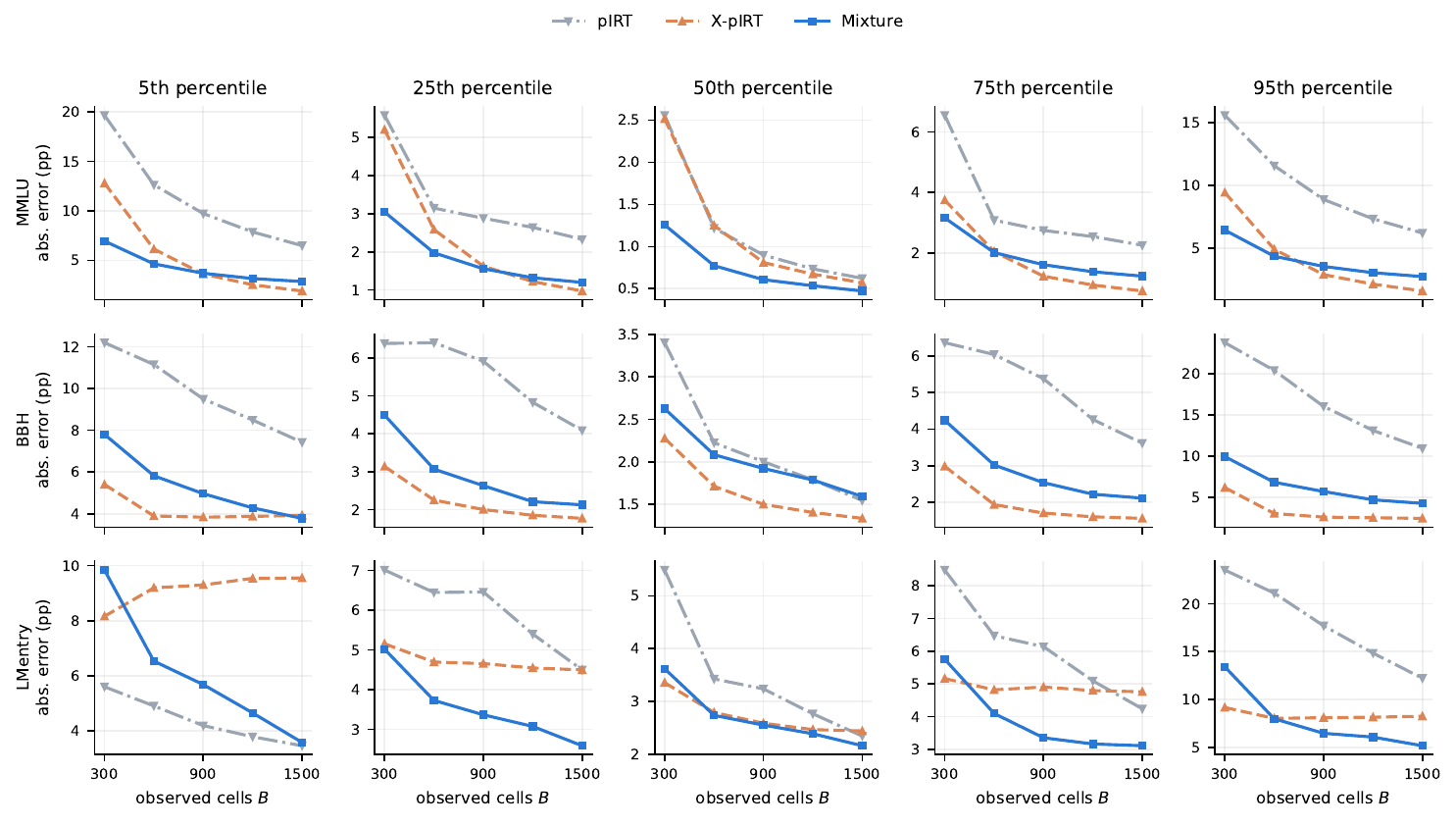}
 \caption{Mixture beats pIRT at almost every quantile, most clearly in the tails. MAE at five quantiles of the prompt score distribution.}
 \label{fig:prompt-quantiles}
\end{figure*}

Table~\ref{tab:prompt-all} reports mean $W_1$ at every budget, and Table~\ref{tab:prompt-gains} gives the reduction relative to pIRT and the share of matrices where Mixture has lower $W_1$. Figure~\ref{fig:prompt-quantiles} reports the absolute error at the 5th, 25th, 50th, 75th, and 95th percentiles of the prompt scores, following PromptEval \citep{polo2024efficient}. Mixture has lower error than pIRT at almost every percentile and budget, with the largest gains at the 5th and 95th percentiles. The exception is the 5th percentile on LMentry, where pIRT is more accurate at every budget. X-pIRT is more accurate than Mixture on BBH at almost every percentile and budget.

\section{BIC component selection}
\label{sec:k-selection}
\begin{figure*}[p]
 \centering
 \includegraphics[width=.8\textwidth]{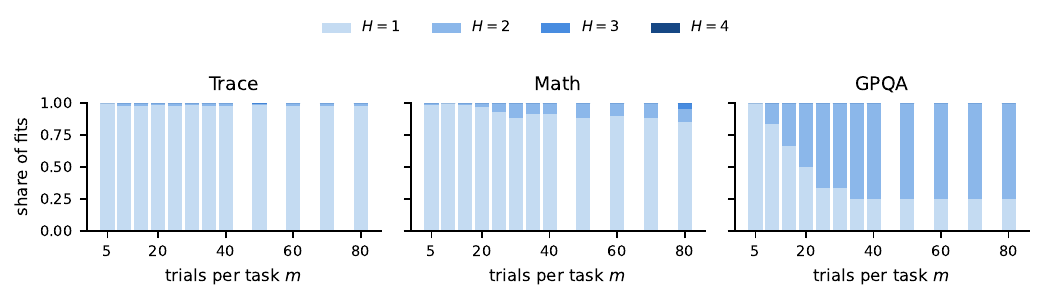}
 \caption{BIC almost always selects $H=1$; GPQA supports a second component as $m$ grows. Share of fits in which BIC selects each $H$.}
 \label{fig:bic-k}
\end{figure*}

BIC selects $H$ separately for every group, seed, and $m$, and the selected fit gives all horizons $k$. Figure~\ref{fig:bic-k} shows how often BIC selects each $H$. Over all fits, it selects $H=1$ in 95\% of cases, with 98\% on Trace, 93\% on Math, and 43\% on GPQA. It never selects $H=4$. On GPQA, every fit at $m=5$ has $H=1$, and from $m=35$ on, 75\% of the fits have $H\geq2$. With $m$ trials per task, the counts depend on $P_\gamma$ only through its first $m$ moments, so the $3H-1$ parameters can be identified only if $3H-1\leq m$. With $m=5$, this allows at most $H=2$. With 3,600 tasks and more trials, the data can support a second component.

\section{Horizon curves by dataset}
\label{sec:dataset-curves}
\begin{figure*}[p]
 \centering
 \includegraphics[width=.85\textwidth]{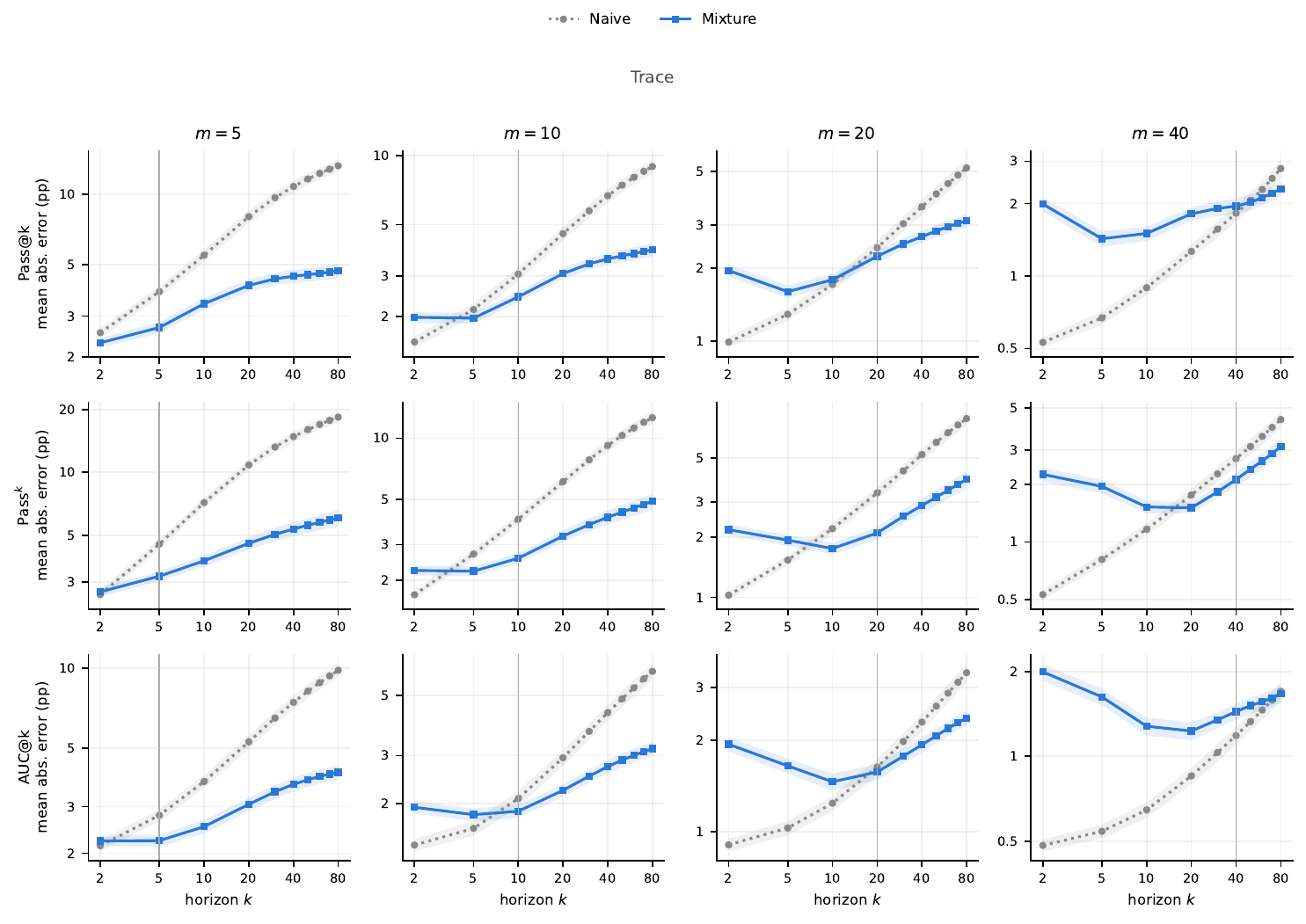}
 \caption{Naive wins for $k$ near or below $m$; Mixture wins as $k$ grows past $m$. Trace MAE as $k$ grows, one column per $m$.}
 \label{fig:trace-k}
\end{figure*}

\begin{figure*}[p]
 \centering
 \includegraphics[width=.85\textwidth]{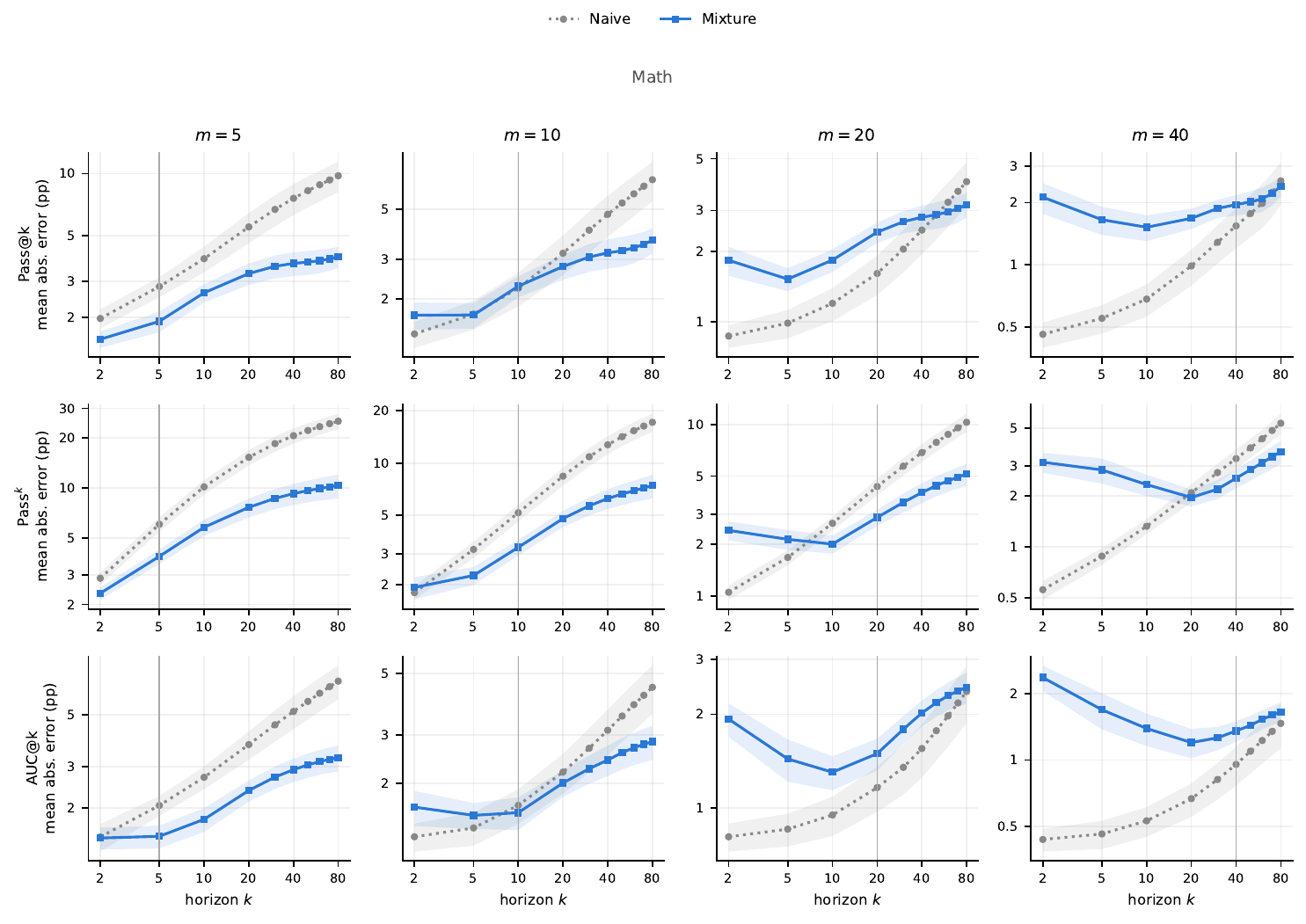}
 \caption{Same pattern as Trace. Same as Figure~\ref{fig:trace-k} for Math.}
 \label{fig:math-k}
\end{figure*}

\begin{figure*}[p]
 \centering
 \includegraphics[width=.85\textwidth]{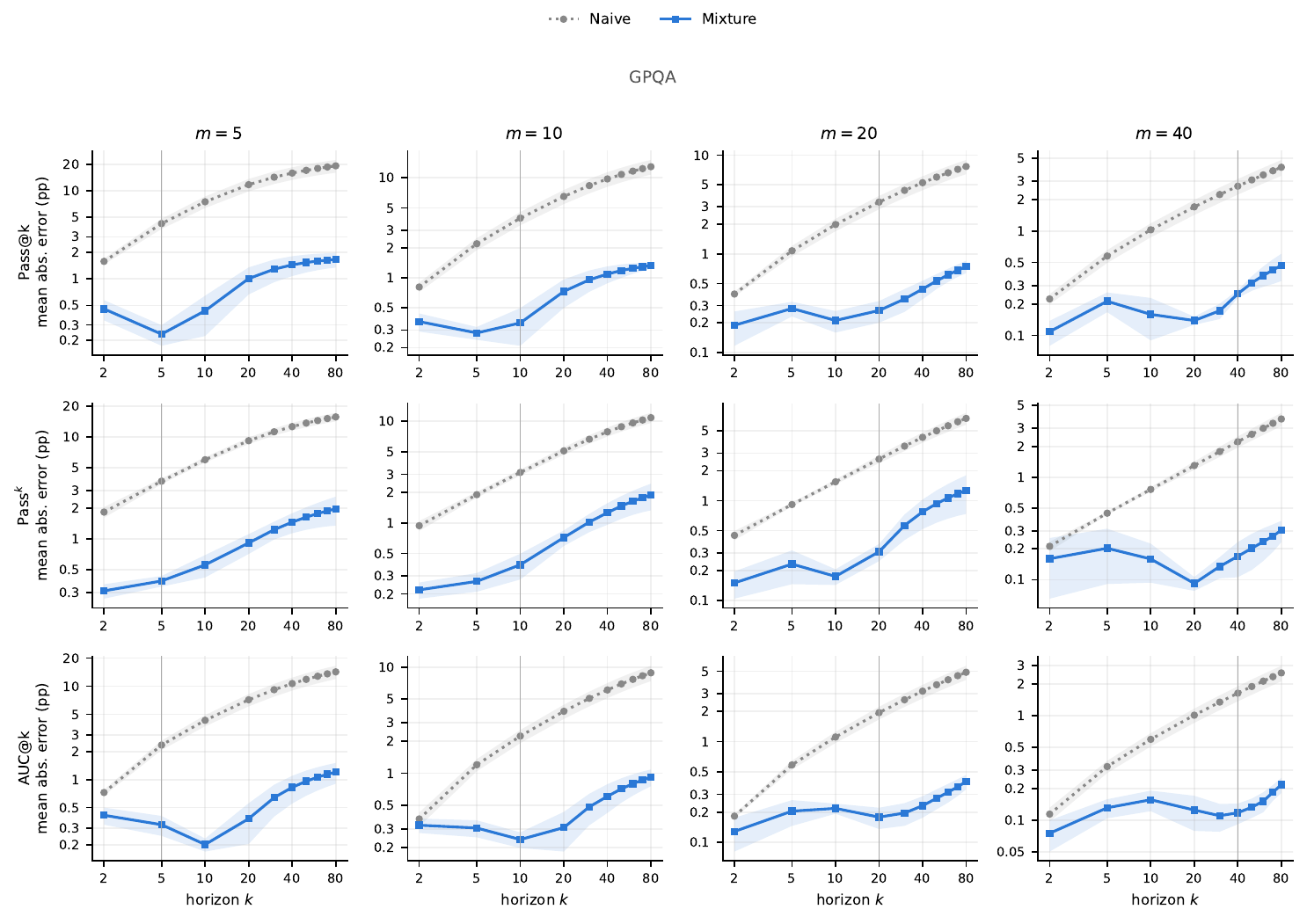}
 \caption{Mixture is more accurate at every $m$ and $k$. Same as Figure~\ref{fig:trace-k} for GPQA.}
 \label{fig:gpqa-k}
\end{figure*}

Figure~\ref{fig:metrics-k} shows only $m=10$ with all groups pooled. Figures~\ref{fig:trace-k}--\ref{fig:gpqa-k} show $m\in\{5,10,20,40\}$ for each dataset. The vertical line marks $k=m$. On Trace and Math, naive is more accurate when $k$ is below or near $m$, and Mixture is more accurate as $k$ grows past $m$. On GPQA, Mixture is more accurate at every $m$ and $k$.

% Larger tables and figures, placed after the appendix text in reference order.
% Figure 6 goes on top of the next page, above Tables 7 and 8.

\newpage
\begin{table}[!htp]
 \centering
 \footnotesize
 \setlength{\tabcolsep}{2.4pt}
 \caption{Combinatorial is the most accurate when $k\leq m$; Mixture is the most accurate when $k>m$. MAE by regime, including the combinatorial estimator.}
 \label{tab:regimes-full}
 \input{tables/regime_full.tex}
\end{table}

\input{tables/k60_by_m.tex}

\clearpage
\begin{table*}[p]
 \centering
 \small
 \caption{Mixture beats naive for every seed, and the seed standard deviation is at most 0.34 points. Extrapolation MAE ($k>m$) for each sampling seed.}
 \label{tab:seed-variation}
 \input{tables/seed_variation.tex}
\end{table*}

\begin{table*}[p]
 \centering
 \small
 \begin{minipage}[t]{.44\textwidth}
 \centering
 \caption{Mixture beats pIRT in every row; X-pIRT, which uses covariates, is best on BBH and on MMLU from $B=900$. Mean $W_1$ error by benchmark and budget.}
 \label{tab:prompt-all}
 \input{tables/prompt_w1.tex}
 \end{minipage}\hfill
 \begin{minipage}[t]{.52\textwidth}
 \centering
 \caption{Mixture beats pIRT on 78--94\% of matrices. Reduction in mean $W_1$ relative to pIRT, and share of matrices where Mixture wins.}
 \label{tab:prompt-gains}
 \setlength{\tabcolsep}{4pt}
 \input{tables/prompt_gains.tex}
 \end{minipage}
\end{table*}

\clearpage

\clearpage

\clearpage

%% file: tables/models.tex
\begin{tabular}{llc}
\toprule
Model & Short name & Collections\\
\midrule
\texttt{AceReason-Nemotron-1.1-7B} & AceReason-7B & Trace\\
\texttt{Bespoke-Stratos-7B} & Bespoke-7B & Trace\\
\texttt{DeepSeek-R1-Distill-Qwen-1.5B} & R1-Distill-1.5B & Trace\\
\texttt{EXAONE-4.0-1.2B} & EXAONE-1.2B & Trace\\
\texttt{FuseO1-DeepSeekR1-QwQ-SkyT1-Flash-32B-Preview} & FuseO1-32B & Trace\\
\texttt{gpt-oss-20b\_high} & gpt-oss-high & Trace, Math, GPQA\\
\texttt{gpt-oss-20b\_low} & gpt-oss-low & Trace, Math, GPQA\\
\texttt{gpt-oss-20b\_medium} & gpt-oss-medium & Trace, Math, GPQA\\
\texttt{Light-R1-14B-DS} & Light-R1-14B & Trace\\
\texttt{LIMO-v2} & LIMO-v2 & Trace\\
\texttt{NVIDIA-Nemotron-Nano-9B-v2} & Nemotron-Nano-9B & Trace\\
\texttt{OpenR1-Distill-7B} & OpenR1-7B & Trace\\
\texttt{OpenReasoning-Nemotron-1.5B} & OR-Nemotron-1.5B & Trace\\
\texttt{OpenThinker2-32B} & OpenThinker2-32B & Trace\\
\texttt{OpenThinker3-1.5B} & OpenThinker3-1.5B & Trace\\
\texttt{Phi-4-reasoning} & Phi-4-R & Trace\\
\texttt{Phi-4-reasoning-plus} & Phi-4-R-plus & Trace\\
\texttt{Qwen3-30B-A3B-Thinking-2507} & Qwen3-30B-A3B & Trace\\
\texttt{Qwen3-4B-Thinking-2507} & Qwen3-4B & Trace\\
\texttt{Qwen3.6-35B-A3B} & Qwen3.6-35B-A3B & Math, GPQA\\
\texttt{Sky-T1-32B-Flash} & Sky-T1-32B & Trace\\
\bottomrule
\end{tabular}

%% file: tables/budget.tex
\begin{tabular}{rrrrrrrrrrrrr}
\toprule
 & \multicolumn{3}{c}{Trace} & \multicolumn{3}{c}{Math} & \multicolumn{3}{c}{GPQA} & \multicolumn{3}{c}{All}\\
Tol. (pp) & Mix. & Naive & Fewer (\%) & Mix. & Naive & Fewer (\%) & Mix. & Naive & Fewer (\%) & Mix. & Naive & Fewer (\%)\\
\midrule
2 & 30 & 50 & 62 & 45 & 60 & 44 & 5 & 55 & 94 & 30 & 50 & 62 \\
3 & 15 & 35 & 81 & 22.5 & 40 & 72 & 5 & 37.5 & 94 & 15 & 35 & 81 \\
4 & 10 & 25 & 88 & 17.5 & 32.5 & 78 & 5 & 27.5 & 94 & 10 & 25 & 88 \\
5 & 5 & 20 & 94 & 10 & 20 & 88 & 5 & 22.5 & 94 & 5 & 20 & 94 \\
6 & 5 & 15 & 94 & 7.5 & 20 & 91 & 5 & 20 & 94 & 5 & 15 & 94 \\
7 & 5 & 15 & 94 & 5 & 15 & 94 & 5 & 15 & 94 & 5 & 15 & 94 \\
\bottomrule
\end{tabular}

%% file: tables/regime_full.tex
\begin{tabular}{llrrrrr}
\toprule
& & \multicolumn{3}{c}{$k\leq m$} & \multicolumn{2}{c}{$k>m$}\\
Dataset & Metric & Naive & Comb. & Mixture & Naive & Mixture\\
\midrule
Trace & Pass@$k$ & 1.19 & 0.88 & 1.72 & 4.89 & 2.91 \\
Trace & Pass$^k$ & 1.58 & 1.03 & 1.85 & 7.07 & 3.51 \\
Trace & AUC@$k$ & 0.83 & 0.65 & 1.46 & 3.24 & 2.21 \\
\addlinespace
Math & Pass@$k$ & 0.95 & 0.68 & 1.68 & 3.70 & 2.79 \\
Math & Pass$^k$ & 1.93 & 1.10 & 2.34 & 9.46 & 4.97 \\
Math & AUC@$k$ & 0.66 & 0.50 & 1.38 & 2.38 & 2.06 \\
\addlinespace
GPQA & Pass@$k$ & 1.42 & 0.10 & 0.20 & 7.18 & 0.70 \\
GPQA & Pass$^k$ & 1.19 & 0.09 & 0.16 & 5.95 & 0.83 \\
GPQA & AUC@$k$ & 0.83 & 0.07 & 0.16 & 4.64 & 0.38 \\
\bottomrule
\end{tabular}

%% file: tables/k60_by_m.tex
\begin{table}[!htp]
\centering
\footnotesize
\caption{Trace: MAE at $k=60$ by $m$. Combinatorial needs $m\geq60$.}\label{tab:k60-trace}
\begin{tabular}{lrrrr}
\toprule
Metric & $m$ & Naive & Comb. & Mixture\\\midrule
Pass@60 & 5 & 12.29 & -- & 4.57 \\
Pass@60 & 10 & 8.03 & -- & 3.75 \\
Pass@60 & 20 & 4.46 & -- & 2.96 \\
Pass@60 & 40 & 2.29 & -- & 2.12 \\
Pass@60 & 60 & 1.61 & 1.16 & 1.69 \\
Pass@60 & 80 & 1.27 & 0.00 & 1.37 \\
\addlinespace
Pass$^{60}$ & 5 & 16.96 & -- & 5.77 \\
Pass$^{60}$ & 10 & 11.21 & -- & 4.54 \\
Pass$^{60}$ & 20 & 6.69 & -- & 3.44 \\
Pass$^{60}$ & 40 & 3.55 & -- & 2.64 \\
Pass$^{60}$ & 60 & 2.44 & 1.46 & 1.97 \\
Pass$^{60}$ & 80 & 1.83 & 0.00 & 1.54 \\
\addlinespace
AUC@60 & 5 & 8.83 & -- & 3.91 \\
AUC@60 & 10 & 5.33 & -- & 3.01 \\
AUC@60 & 20 & 2.87 & -- & 2.19 \\
AUC@60 & 40 & 1.46 & -- & 1.56 \\
AUC@60 & 60 & 0.94 & 0.70 & 1.30 \\
AUC@60 & 80 & 0.74 & 0.00 & 1.00 \\
\bottomrule
\end{tabular}
\end{table}

\begin{table}[!htp]
\centering
\footnotesize
\caption{Math: MAE at $k=60$ by $m$. Combinatorial needs $m\geq60$.}\label{tab:k60-math}
\begin{tabular}{lrrrr}
\toprule
Metric & $m$ & Naive & Comb. & Mixture\\\midrule
Pass@60 & 5 & 8.82 & -- & 3.77 \\
Pass@60 & 10 & 5.85 & -- & 3.38 \\
Pass@60 & 20 & 3.25 & -- & 2.96 \\
Pass@60 & 40 & 1.98 & -- & 2.08 \\
Pass@60 & 60 & 1.35 & 0.92 & 1.82 \\
Pass@60 & 80 & 1.07 & 0.00 & 1.45 \\
\addlinespace
Pass$^{60}$ & 5 & 23.35 & -- & 9.94 \\
Pass$^{60}$ & 10 & 15.36 & -- & 6.94 \\
Pass$^{60}$ & 20 & 8.76 & -- & 4.69 \\
Pass$^{60}$ & 40 & 4.34 & -- & 3.13 \\
Pass$^{60}$ & 60 & 2.99 & 1.41 & 2.56 \\
Pass$^{60}$ & 80 & 2.35 & 0.00 & 1.87 \\
\addlinespace
AUC@60 & 5 & 6.19 & -- & 3.15 \\
AUC@60 & 10 & 3.86 & -- & 2.69 \\
AUC@60 & 20 & 1.98 & -- & 2.29 \\
AUC@60 & 40 & 1.23 & -- & 1.53 \\
AUC@60 & 60 & 0.72 & 0.56 & 1.27 \\
AUC@60 & 80 & 0.58 & 0.00 & 0.89 \\
\bottomrule
\end{tabular}
\end{table}

\begin{table}[!htp]
\centering
\footnotesize
\caption{GPQA: MAE at $k=60$ by $m$. Combinatorial needs $m\geq60$.}\label{tab:k60-gpqa}
\begin{tabular}{lrrrr}
\toprule
Metric & $m$ & Naive & Comb. & Mixture\\\midrule
Pass@60 & 5 & 17.93 & -- & 1.59 \\
Pass@60 & 10 & 11.67 & -- & 1.24 \\
Pass@60 & 20 & 6.64 & -- & 0.61 \\
Pass@60 & 40 & 3.45 & -- & 0.38 \\
Pass@60 & 60 & 2.32 & 0.09 & 0.20 \\
Pass@60 & 80 & 1.74 & 0.00 & 0.13 \\
\addlinespace
Pass$^{60}$ & 5 & 14.49 & -- & 1.78 \\
Pass$^{60}$ & 10 & 9.63 & -- & 1.63 \\
Pass$^{60}$ & 20 & 5.64 & -- & 1.07 \\
Pass$^{60}$ & 40 & 3.00 & -- & 0.23 \\
Pass$^{60}$ & 60 & 2.08 & 0.13 & 0.19 \\
Pass$^{60}$ & 80 & 1.59 & 0.00 & 0.11 \\
\addlinespace
AUC@60 & 5 & 12.81 & -- & 1.06 \\
AUC@60 & 10 & 7.68 & -- & 0.80 \\
AUC@60 & 20 & 4.13 & -- & 0.31 \\
AUC@60 & 40 & 2.12 & -- & 0.15 \\
AUC@60 & 60 & 1.41 & 0.07 & 0.07 \\
AUC@60 & 80 & 1.06 & 0.00 & 0.09 \\
\bottomrule
\end{tabular}
\end{table}

%% file: tables/seed_variation.tex
\begin{tabular}{lllrrrrr}
\toprule
Dataset & Metric & Estimator & Seed 0 & Seed 1 & Seed 2 & Mean & SD\\
\midrule
Trace & Pass@$k$ & Naive & 4.96 & 5.09 & 4.63 & 4.89 & 0.24 \\
Trace & Pass@$k$ & Mixture & 2.93 & 2.77 & 3.01 & 2.91 & 0.12 \\
Trace & Pass$^k$ & Naive & 6.79 & 6.99 & 7.44 & 7.07 & 0.34 \\
Trace & Pass$^k$ & Mixture & 3.34 & 3.55 & 3.64 & 3.51 & 0.15 \\
Trace & AUC@$k$ & Naive & 3.31 & 3.40 & 3.03 & 3.24 & 0.19 \\
Trace & AUC@$k$ & Mixture & 2.24 & 2.11 & 2.27 & 2.21 & 0.08 \\
\addlinespace
Math & Pass@$k$ & Naive & 3.50 & 3.94 & 3.65 & 3.70 & 0.22 \\
Math & Pass@$k$ & Mixture & 2.72 & 2.74 & 2.92 & 2.79 & 0.11 \\
Math & Pass$^k$ & Naive & 9.71 & 9.25 & 9.43 & 9.46 & 0.23 \\
Math & Pass$^k$ & Mixture & 4.99 & 4.86 & 5.04 & 4.97 & 0.09 \\
Math & AUC@$k$ & Naive & 2.27 & 2.55 & 2.34 & 2.38 & 0.14 \\
Math & AUC@$k$ & Mixture & 1.97 & 1.94 & 2.27 & 2.06 & 0.18 \\
\addlinespace
GPQA & Pass@$k$ & Naive & 7.17 & 7.07 & 7.30 & 7.18 & 0.12 \\
GPQA & Pass@$k$ & Mixture & 0.59 & 0.71 & 0.79 & 0.70 & 0.10 \\
GPQA & Pass$^k$ & Naive & 5.86 & 6.08 & 5.92 & 5.95 & 0.12 \\
GPQA & Pass$^k$ & Mixture & 0.83 & 0.76 & 0.88 & 0.83 & 0.06 \\
GPQA & AUC@$k$ & Naive & 4.63 & 4.55 & 4.72 & 4.64 & 0.09 \\
GPQA & AUC@$k$ & Mixture & 0.32 & 0.37 & 0.45 & 0.38 & 0.07 \\
\bottomrule
\end{tabular}

%% file: tables/prompt_w1.tex
\begin{tabular}{lrrrr}
\toprule
Benchmark & $B$ & pIRT & X-pIRT & Mixture\\\midrule
MMLU & 300 & 8.53 & 5.90 & 3.68 \\
MMLU & 600 & 5.31 & 2.99 & 2.42 \\
MMLU & 900 & 4.30 & 1.81 & 1.94 \\
MMLU & 1200 & 3.66 & 1.34 & 1.66 \\
MMLU & 1500 & 3.11 & 1.05 & 1.49 \\
\addlinespace
BBH & 300 & 8.67 & 3.70 & 5.25 \\
BBH & 600 & 7.78 & 2.43 & 3.80 \\
BBH & 900 & 6.73 & 2.20 & 3.28 \\
BBH & 1200 & 5.59 & 2.11 & 2.83 \\
BBH & 1500 & 4.76 & 2.06 & 2.60 \\
\addlinespace
LMentry & 300 & 8.91 & 5.51 & 6.58 \\
LMentry & 600 & 7.44 & 5.14 & 4.48 \\
LMentry & 900 & 6.82 & 5.12 & 3.88 \\
LMentry & 1200 & 5.75 & 5.09 & 3.62 \\
LMentry & 1500 & 4.84 & 5.08 & 3.20 \\
\bottomrule
\end{tabular}

%% file: tables/prompt_gains.tex
\begin{tabular}{lrrrr}
\toprule
& \multicolumn{2}{c}{$W_1$ gain (pp)} & \multicolumn{2}{c}{Win share (\%)}\\
Benchmark & $B=300$ & $B=1500$ & $B=300$ & $B=1500$\\
\midrule
MMLU & 4.85 & 1.62 & 93 & 94 \\
BBH & 3.42 & 2.16 & 88 & 84 \\
LMentry & 2.33 & 1.64 & 81 & 78 \\
\bottomrule
\end{tabular}